\documentclass[sigconf,nonacm]{acmart}

\usepackage[utf8]{inputenc}
\usepackage[T1]{fontenc}

\usepackage{microtype}
\usepackage{mdframed}      
\usepackage{xcolor}
\usepackage{lipsum}
\usepackage{enumitem}
\usepackage{booktabs}
\usepackage{caption}
\usepackage{todonotes}
\usepackage[normalem]{ulem}
\useunder{\uline}{\ul}{}
\usepackage{xspace}
\usepackage{adjustbox}
\usepackage{tcolorbox}

\newmdenv[backgroundcolor=gray!30]{grayparagraph}
\newcommand{\dsname}{\textit{Poly-FEVER}\xspace}

\setcopyright{none}

\begin{document}

\title{\dsname: A Multilingual Fact Verification Benchmark for Hallucination Detection in Large Language Models}

\author{Hanzhi Zhang}
\affiliation{%
  \institution{University of North Texas}
  \city{Denton}
  \state{TX}
  \country{United States}
}
\email{hanzhizhang@my.unt.edu}

\author{Sumera Anjum}
\affiliation{%
  \institution{University of North Texas}
  \city{Denton}
  \state{TX}
  \country{United States}
}
\email{sumeraanjumsumeraanjum@my.unt.edu}

\author{Heng Fan}
\affiliation{%
  \institution{University of North Texas}
  \city{Denton}
  \state{TX}
  \country{United States}
}
\email{heng.fan@unt.edu}

\author{Weijian Zheng}
\affiliation{%
  \institution{Argonne National Laboratory}
  \city{Lemont}
  \state{IL}
  \country{United States}
}
\email{wzheng@anl.gov}

\author{Yan Huang}
\affiliation{%
  \institution{University of North Texas}
  \city{Denton}
  \state{TX}
  \country{United States}
}
\email{yan.huang@unt.edu}

\author{Yunhe Feng}
\affiliation{%
  \institution{University of North Texas}
  \city{Denton}
  \state{TX}
  \country{United States}
}
\email{yunhe.feng@unt.edu}

\begin{abstract}

Hallucinations in generative AI, particularly in Large Language Models (LLMs), pose a significant challenge to the reliability of multilingual applications. Existing benchmarks for hallucination detection focus primarily on English and a few widely spoken languages, lacking the breadth to assess inconsistencies in model performance across diverse linguistic contexts. To address this gap, we introduce \dsname, a large-scale multilingual fact verification benchmark specifically designed for evaluating hallucination detection in LLMs. \dsname comprises 77,973 labeled factual claims spanning 11 languages, sourced from FEVER, Climate-FEVER, and SciFact. It provides the first large-scale dataset tailored for analyzing hallucination patterns across languages, enabling systematic evaluation of LLMs such as ChatGPT and the LLaMA series. Our analysis reveals how topic distribution and web resource availability influence hallucination frequency, uncovering language-specific biases that impact model accuracy. By offering a multilingual benchmark for fact verification, \dsname facilitates cross-linguistic comparisons of hallucination detection and contributes to the development of more reliable, language-inclusive AI systems. The dataset is publicly available to advance research in responsible AI, fact-checking methodologies, and multilingual NLP, promoting greater transparency and robustness in LLM performance. The proposed \dsname is available at: \textcolor{blue}{\url{https://huggingface.co/datasets/HanzhiZhang/Poly-FEVER}}.

\end{abstract}

\maketitle

\section{Introduction}

LLMs, such as those in the GPT family, have exhibited remarkable proficiency across diverse domains including education, healthcare, and legal affairs. In these applications, the accuracy and factual integrity of the content generated by LLMs are critical, particularly in areas requiring precise guidance, like medical and legal advice. Despite their advancements, mainstream LLMs predominantly utilize corpora that are imbalanced in terms of demographic groups~\cite{shah2019predictive,li2023survey}. 
Language, an important facet of demographic backgrounds, remains relatively underinvestigated in the context of detecting hallucinatory content in LLMs, particularly from the perspective of fairness and equitable usability.

Previous research has predominantly focused on hallucinations in LLMs within widely spoken languages, such as English~\cite{yao2023llm}, Chinese~\cite{cheng2023evaluating}, and German~\cite{sennrich2023mitigating}. This focus has led to a thorough understanding of hallucinatory outputs’ mechanisms and the development of mitigation strategies. 
One approach involves prompt engineering, which includes retrieval augmentation to ground content in external evidence~\cite{lewis2020retrieval}, feedback loops for refining responses, and prompt tuning to adjust prompts during fine-tuning for desired behaviors. 
Another strategy is model development~\cite{tonmoy2024comprehensive}, focusing on creating models inherently less prone to hallucinating through architectural changes, novel loss functions, and supervised fine-tuning using human-labeled data.

Despite the critical insights gained, the focus on major languages has marginalized the experiences and challenges of LLMs trained on or applied to less common languages. 
Moreover, these investigations often employ differing datasets, leading to an absence of a systematic approach to assessing hallucinations across languages with uniform input. 
The complexity of this issue is multifold. 
First, there is a scarcity of appropriate datasets for cross-linguistic studies. Second, accurately detecting hallucinations on a large scale can be challenging, particularly in topics intertwined with local cultural and linguistic contexts. Third, the generation of apparently plausible yet inaccurate content manifests due to biased training and fine-tuning referenced resources on languages of marginal transmission.
Nonetheless, ensuring equitable performance of LLMs across languages, especially those that are underrepresented, is crucial from the perspective of hallucination consistency.

This paper examines hallucination detection in LLMs across multiple languages by expanding fact verification into a multilingual benchmark, \dsname. Fact verification and hallucination detection involve different objectives. Fact verification determines whether a claim aligns with established knowledge sources \cite{murayama2021dataset, zhu2021knowledge}. This process requires external evidence to confirm or refute statements. Hallucination detection, however, identifies false or misleading claims without relying on external references. An LLM that fabricates information may generate statements that appear factual but contain errors, inconsistencies, or contradictions. Evaluating hallucinations requires analyzing the model’s outputs to detect inaccuracies, regardless of whether external knowledge exists. In multilingual contexts, hallucination detection becomes more complex because models process languages with different data distributions, resource availability, and linguistic structures. \dsname addresses this challenge by enabling a structured evaluation of hallucinations across multiple languages, exposing discrepancies in model performance and assessing factual consistency without dependence on external sources.

We summarize the key contributions as follows: 
\begin{enumerate}
    \item We introduce {\dsname}, an extensive, publicly available dataset tailored for multilingual fact extraction and verification. It covers 11 languages and includes 77,973 fact claims on various topics, designed for hallucination detection tasks. \dsname is available at: \textcolor{blue}{\url{https://huggingface.co/datasets/HanzhiZhang/Poly-FEVER}}.
    \item We analyze hallucination detection capabilities in advanced language models of ChatGPT, LLaMA-2 series (7B, 13B, and 70B), and LLaMA-3.1 8B, using {\dsname} with both language-wise and classification prompts.
    \item We investigate the reasons behind hallucinations on a multilingual scale, employing Latent Dirichlet Allocation (LDA) for topic distribution analysis and automated web searches to assess resource imbalances.
\end{enumerate}

\section{Related Work}

LLMs generate hallucinations—outputs that contradict known facts or fabricate non-existent information, posing challenges for reliable AI applications. \citet{huang2023survey} categorize hallucinations into two types: intrinsic and extrinsic. Intrinsic hallucinations occur when a model produces self-contradictions or logical inconsistencies within its own response, while extrinsic hallucinations involve factually incorrect statements that do not align with established knowledge sources. Evaluating hallucinations requires a systematic comparison between generated content and verified ground-truth data or an analysis of internal inconsistencies within model outputs. Effective hallucination detection plays a crucial role in ensuring the credibility of generative AI systems, particularly in high-stakes applications such as healthcare, legal services, and scientific research.

Existing evaluation metrics measure hallucination frequency by assessing entity correctness, factual consistency, and contradiction detection. Early methods relied on n-gram overlap (e.g., ROUGE, PARENT-T) \cite{lin-2004-rouge, wang-etal-2020-towards}, but these metrics fail to capture factual accuracy. Entity-based hallucination metrics \cite{nan-etal-2021-entity} and relation-based fact extraction models \cite{10.1145/3292500.3330955} provide more structured approaches. However, these methods primarily target English-language outputs, limiting their applicability to multilingual hallucination analysis.

To assess hallucination detection, researchers rely on fact-checking datasets such as FEVER \cite{thorne2018fever}, Climate-FEVER \cite{diggelmann2020climate}, and SciFact \cite{wadden2020fact}. These datasets evaluate model responses by comparing claims against external sources. However, they primarily focus on fact verification, not hallucination detection, since they assume that relevant evidence is always available.

Current benchmarks also exhibit strong language biases. Most fact-checking datasets contain English-only claims, restricting their ability to evaluate how LLMs handle fact verification across different linguistic contexts. While some efforts attempt multilingual fact-checking \cite{gupta-srikumar-2021-x}, these datasets typically do not provide identical claims across languages, making it difficult to access cross-lingual hallucination patterns.

Most research on multilingual hallucinations focuses on machine translation errors, where LLMs generate hallucinated translations \cite{pfeiffer2023mmt5}. However, translation-based hallucination benchmarks do not evaluate self-contained factual inconsistencies within LLM outputs.

LLMs trained on English-centric corpora often exhibit language-specific hallucination patterns, but existing datasets do not provide a systematic, multilingual benchmark for evaluating hallucination detection across languages. A benchmark that contains the same factual claims across multiple languages, evaluates hallucination detection directly, and exposes cross-linguistic biases is missing from current research.

Poly-FEVER fills this gap by introducing a large-scale multilingual benchmark for hallucination detection in LLMs. It extends FEVER, Climate-FEVER, and SciFact to 11 languages, covering 77,973 factual claims. Unlike previous datasets, Poly-FEVER does not assume external evidence is always available, making it suitable for testing LLMs' intrinsic hallucination detection abilities in multilingual settings.

\section{{\dsname} Benchmark}

This section details the construction, structure, and application of the {\dsname} benchmark. We describe the dataset curation process, present its organizational structure, analyze the distribution of topics within the data, and evaluate hallucination detection across languages using a range of LLMs.

\subsection{{\dsname} Overview}

{\dsname} provides a multilingual benchmark for fact verification and hallucination detection in large language models (LLMs). It contains 77,973 labeled claims in 11 languages. The dataset extends three widely used English fact-checking sources: FEVER (Fact Extraction and VERification)~\cite{thorne2018fever}, Climate-FEVER~\cite{diggelmann2020climate}, and SciFact~\cite{wadden2020fact}. These sources establish a structured evaluation framework for fact verification by defining claims and classifying them based on supporting or refuting evidence.

FEVER contains 185,445 claims derived from Wikipedia content. The dataset assigns each claim one of three labels: \textit{Supported}, \textit{Refuted}, or \textit{NotEnoughInfo}. The first two categories indicate whether verifiable evidence exists, while the third highlights instances where information remains insufficient. {\dsname} excludes claims labeled as \textit{NotEnoughInfo} to focus on cases where factual accuracy can be measured directly.

Climate-FEVER builds on the FEVER framework but focuses on climate-related claims. The dataset includes statements verified against scientific literature and expert-reviewed sources. It provides a structured evaluation of fact verification in a domain that requires specialized knowledge. {\dsname} incorporates only \textit{Supported} and \textit{Refuted} claims from Climate-FEVER to maintain consistency across sources.

SciFact contains scientific claims extracted from biomedical literature. The dataset evaluates whether scientific papers \textit{Support} or \textit{Refute} a given claim. SciFact ensures that the benchmark covers technical content that requires precise verification. By including these claims, {\dsname} enables a structured evaluation of hallucinations in scientific domains.

The dataset spans diverse topics, including Arts, Music, Science, Biology, and History, ensuring comprehensive fact verification across a wide range of subject areas. These topics vary in specificity and contextual complexity, allowing for an in-depth analysis of how LLMs handle factual claims across different domains. The multilingual nature of {\dsname} further adds to its complexity, as it introduces variations in linguistic structure, requiring models to process and verify claims across languages with distinct grammar, syntax, and lexical properties. By incorporating multiple languages, {\dsname} facilitates a more nuanced evaluation of LLMs’ ability to distinguish between factual and hallucinatory content across linguistic and cultural boundaries. Through its structured and diverse dataset, {\dsname} supports research in hallucination detection by providing a benchmark that enables systematic assessment of factual consistency, model biases, and the effectiveness of mitigation strategies across different linguistic contexts.

\subsection{Structure of {\dsname}}

The {\dsname} dataset follows the structure of the FEVER family of datasets (FEVER, Climate-FEVER, and SciFact), facilitating direct comparisons and leveraging established fact-verification methodologies. Each entry in {\dsname} contains four primary fields:

\begin{itemize}[leftmargin=10pt]
    \item \textbf{ID:} A unique identifier assigned to each claim, allowing for consistent referencing and seamless integration with other annotations or future expansions.
    \item \textbf{Label:} An annotated veracity label (\textit{true} or \textit{false}) indicating whether the claim aligns with established factual evidence. This ensures that each claim is directly tied to a definitive truth value, reflecting its correctness within the dataset.
    \item \textbf{Topic Distribution:} A vector of the top five predicted topics for the claim, derived through thematic analysis (e.g., LDA). This field offers contextual clues about the claim’s domain (e.g., sports, history, science) and aids in cross-topic assessments of model performance.
    \item \textbf{Claim:} The textual content of the claim, is provided in 11 distinct languages. This multilingual format highlights the dataset’s cross-lingual capabilities, enabling evaluations of how models handle claims across diverse linguistic and cultural contexts.
\end{itemize}

This four-field structure, especially the multilingual claim component, supports robust assessments of fact-verification methods in varied linguistic settings. It aligns with our core research aim: evaluating language model performance and hallucination tendencies when tasked with fact-checking across different topics, domains, and languages.

\subsection{LDA on Topic Distribution \label{sec:LDA_intro}}

To analyze hallucination patterns in multilingual fact verification, we apply Latent Dirichlet Allocation (LDA)~\cite{blei2003latent} to identify thematic structures in claims. The model evaluates 22 topics in the Table~\ref{tab:num_topic_check} across 11 languages, examining how topic distribution relates to hallucination frequency.

LDA models each claim as a mixture of topics, with each topic represented by a distribution of words. The model assumes that claims share underlying themes and assigns probability distributions to capture these relationships. Each claim consists of weighted topic assignments that indicate its alignment with different themes. The model learns these distributions by adjusting word probabilities to match patterns observed in the dataset. We evaluate topic classifications from 0 to 50 and determine that 22 topics achieve the highest coherence scores on {\dsname}.

To prepare claims for topic modeling, we apply a preprocessing pipeline that standardizes text for analysis. The pipeline converts all text to lowercase, tokenizes words, removes typos, eliminates stopwords, and applies stemming and lemmatization. This process ensures that LDA captures relevant word distributions without noise from text inconsistencies.

After preprocessing, we construct a Gensim Dictionary and transform the corpus into a Bag-of-Words (BoW) representation. We apply Term Frequency-Inverse Document Frequency (TF-IDF) weighting to refine feature representation. The LDA model runs for 200 iterations to identify topic distributions and optimize coherence scores. The model assigns probabilities to claims based on topic composition, revealing structures that influence hallucination patterns in multilingual verification. Section~\ref{sec:LDA_exp} provides a detailed analysis of topic distributions and their impact on hallucination detection.

\begin{table}[htbp]
\centering
\setlength{\tabcolsep}{5pt} 
\begin{tabular}{cl|cl}
\toprule
\textbf{ID} & \textbf{Topic} & \textbf{ID} & \textbf{Topic} \\
\midrule
0  & Politics                 & 11 & Equestrian                    \\
1  & Music                    & 12 & Architecture/Construction     \\
2  & Sports                   & 13 & Entertainment                 \\
3  & Aviation                 & 14 & Automotive Racing             \\
4  & Competitive Sports       & 15 & Education                     \\
5  & American Football        & 16 & Soccer                        \\
6  & Photography/Arts         & 17 & Biology/Taxonomy              \\
7  & Demographics             & 18 & Architecture/History          \\
8  & Warfare History          & 19 & Business/Finance              \\
9  & Military Services        & 20 & Film/Television               \\
10 & Biology/Genetics         & 21 & Transportation/Infrastructure \\
\bottomrule
\end{tabular}
\caption{Number-topic references table for LDA topic distribution on 22 topics. This table presents the topic assignments derived from LDA applied to the {\dsname} dataset. These topic classifications serve as a basis for evaluating the impact of topic awareness.}
\vspace{-1cm}
\label{tab:num_topic_check}
\end{table}

\subsection{Language Selection}

{\dsname} includes claims in 11 languages to analyze language bias in hallucination detection. The dataset covers English (en), Mandarin Chinese (zh-CN), Hindi (hi), Arabic (ar), Bengali (bn), Japanese (ja), Korean (ko), Tamil (ta), Thai (th), Georgian (ka), and Amharic (am). These languages appear in order of the number of native speakers. The selection ensures that the benchmark evaluates LLM performance across linguistic groups with different structures, resources, and usage patterns.

Each language presents challenges that influence hallucination patterns in LLMs. Mandarin Chinese, Arabic, and Japanese include logographic or syllabic writing systems that differ from English's alphabetic structure. Tamil, Thai, and Georgian introduce distinct syntactic rules that affect sentence formation. Ambiguity, polysemy, and homophones in Hindi, Korean, and Bengali require models to resolve multiple interpretations of words in context. Amharic and Tamil contain complex morphological patterns that affect tokenization and semantic understanding.

Dialects and sociolects create additional challenges in fact verification. Hindi and Arabic contain regional variations that influence word choice and sentence construction. Korean and Japanese incorporate honorifics and context-dependent expressions that require models to infer meaning from cultural and social cues. LLMs trained primarily on English and widely spoken languages struggle to adapt to these variations, leading to inconsistencies in hallucination detection.

The selection also considers differences in script and orthography. Non-Latin scripts appear in Chinese, Arabic, Thai, Hindi, Bengali, Tamil, Georgian, and Amharic. These scripts require different tokenization strategies, which affect how models encode and process text. The dataset ensures that evaluations reflect performance across writing systems with unique linguistic properties.

\subsection{Multilingual Claim Translation}

Google Cloud Translation extends the benchmark to 11 languages. The translation process required evaluating different methods, including translation APIs like DeepL Translation and LLM-based translation models. Each method was tested to determine its accuracy in preserving meaning across languages. 

Several coauthors fluent in multiple languages assessed translation quality by reviewing selected claims in two languages. The evaluation focused on preserving cultural and contextual meaning. Google Translate produced more accurate translations than other methods. DeepL Translate introduced structural errors, such as altering verb tense and misordering words. One example included the mistranslation of "The book was read by him" as "Book by him read," which changed the grammatical structure. These inconsistencies led to the exclusion of DeepL and other APIs.

Concerns about hallucinations in LLM-based translations influenced the decision to use Google Translate. LLMs generate fluent text but may introduce fabricated details when translating factual claims. Google Cloud Translation provided greater consistency by following source structures without generating new information. 

The translation process covered more than 80,000 English factual claims across 10 languages. The total cost for Google Cloud Translation reached \$2,644 USD. To evaluate translation quality, the GPT Estimation Metric Based Assessment (GEMBA)~\cite{kocmi-federmann-2023-large} analyzed 5\% of the benchmark. GEMBA does not require human reference translations and provides a quantitative assessment of meaning preservation. Table~\ref{tab:trans_eva} presents the evaluation results.

\begin{table}[htbp]
\centering
\setlength{\tabcolsep}{2.5pt} 
\begin{tabular}{ccccccccccc}
\toprule
\textbf{Lang.}    & \textbf{zh} & \textbf{hi} & \textbf{ar} & \textbf{bn} & \textbf{ja} & \textbf{ko}    & \textbf{ta} & \textbf{th} & \textbf{ka} & \textbf{am} \\
\midrule
\textbf{AveScore} & 91.3          & 92.4       & 90.8       & 91.8       & 91.5       & \textbf{93.0} & 90.0          & 91.1       & 90.8       & {\ul 88.9} \\
\bottomrule
\end{tabular}
\caption{Average scores for each language on translation quality evaluation of 5\% {\dsname} benchmark. A score of zero means no meaning preserved and a score of one hundred means perfect meaning and grammar.}
\vspace{-0.8cm}
\label{tab:trans_eva}
\end{table}

\section{Hallucination Causes Exploration}

Hallucinations in LLMs vary across languages and tasks, influenced by model training, available references, and prompt design. To understand these variations, we examine hallucination detection through three perspectives. First, we evaluate how models classify factual claims under different prompts and whether they recognize their own hallucinations. Second, we analyze external factors, such as the availability of references on the web, which may contribute to hallucination rates. Third, we assess whether retrieval-augmented generation (RAG) improves factual accuracy by providing external evidence. Together, these analyses reveal key factors affecting hallucination patterns in multilingual fact verification.

\subsection{Model Performance}

\subsubsection{Effect of Prompt Design}

LLMs often struggle to classify factual claims accurately, as their responses may include irrelevant explanations, hesitations, or references instead of clear binary answers. Effective prompt design plays a crucial role in guiding models toward structured outputs. To enforce strict response constraints, we introduce a set of prompts tailored to ensure consistency across languages and models. These include a general prompt, a language-specific prompt, and a classification prompt, each designed to refine hallucination detection.

We designed both a language-wise prompt and a classification prompt (each with 11 translated versions) to guide LLMs toward producing either \textit{true} or \textit{false} outputs for fact-checking tasks. During initial trials, we noticed that complex, information-rich prompts often confused the models, causing them to overlook specific instructions and generate irrelevant responses. To address this, we iteratively refined our prompt text, focusing on clarity and brevity. This adjustment ensured that the prompts conveyed the essential task constraints without overwhelming the model. Ultimately, we developed a concise general prompt to classify factual claims as \textit{true} or \textit{false} based on truthfulness, thereby reducing irrelevant explanations and improving the accuracy of LLM responses.

\begin{tcolorbox}[colback=gray!10, boxrule=1pt, colframe=black, title=General Prompt]
Label the following claim with truthfulness into: \textit{true} and \textit{false}. No other characters, no punctuation, all lower size.
\end{tcolorbox}

Nevertheless, LLMs often exhibit overenthusiasm by providing detailed explanations or references, straying from the intended binary classification of each claim as \textit{true} or \textit{false}. To counter this, we restructured the language-wise prompt into a three-sentence directive outlining the overall task definition, the precise answer format, and forbidden responses. We observed that model performance deteriorates when prompts and claims diverge linguistically, so we translated this standardized prompt text into the same 11 languages used for the claims. This alignment between prompt language and claim language aims to curb irrelevant model outputs and ensure consistency in binary fact-checking responses.

\begin{tcolorbox}[colback=gray!10, boxrule=1pt, colframe=black, title=Language-wise Prompt]
Assess the truthfulness of the user's claim and provide a response. Use \textit{true} to indicate that the claim is true, \textit{false} to indicate that it is false. Your response should only consist of \textit{true} or \textit{false}, without any additional characters or punctuation.
\end{tcolorbox}

Although LLMs are directed to evaluate claims as \textit{true} or \textit{false} without contextual analysis, their responses are integrated sentence structure. Concerned with classification accuracy, we abandoned simple traditional models to understand the contextual preference of \textit{true} or \textit{false} in LLM responses. The classification prompt guides LLMs to categorize responses. For consistency and to control for hallucinations, responses are translated into English before classification, ensuring binary labels are clear for subsequent analysis.

\begin{tcolorbox}[colback=gray!10, boxrule=1pt, colframe=black, title=Classification Prompt]
Classify the input as \textit{true} or \textit{false} based solely on the indicative words or phrases within it. Use true for it contains affirming words like \textit{Correct,} \textit{TRUE,} \textit{really,} or \textit{the truth.} Use \textit{false} for it contains negating or contradictory phrases like \textit{Fake,} \textit{False,} or any form of correction or contradiction within the input. Respond with only \textit{true} or \textit{false} for the input, without any additional text, characters, or punctuation. 
\end{tcolorbox}

By filtering responses through this structured classification prompt, we ensure that outputs conform to a strict binary format. This approach standardizes hallucination detection across multiple languages, preventing inconsistencies in model interpretation.

\subsubsection{Self-Detection of Hallucinations in LLMs}

We address a concern regarding the capacity of LLMs to identify and mitigate hallucinations in the text they generate. Despite the construction of a {\dsname} dataset aimed at detecting hallucinations, a gap exists in the literature concerning the effectiveness of these models in recognizing inaccuracies within their own outputs. This gap stems from the fact that the claims and labels in {\dsname} are not produced by LLMs, raising questions about the representativeness of the hallucinations that LLMs themselves produce.

\begin{tcolorbox}[colback=gray!10, boxrule=1pt, colframe=black, title=Rephrase Prompt]
Rephrase the following claim without changing its meaning. Ensure the essence and intent remain unchanged.
\end{tcolorbox}

To bridge this gap, we instruct LLMs to rephrase dataset claims in multiple languages while keeping their original meaning intact. 
Claims generated by LLMs lack verifiable ground truth, which is essential for systematically assessing the model's hallucination detection accuracy. 
Given this limitation, we rephrased 77,973 claims to simulate LLM-generated content while preserving each claim's ground truth. 
This controlled evaluation offers a consistent baseline for comparing the performance of spontaneously generated claims without ground truth. 
Due to the lack of automated metrics, we randomly selected 100 claims in English and Chinese and found that their truthfulness remained unchanged.
By doing this, we aim to evaluate whether LLMs can effectively detect hallucinations in their own generated text with similar accuracy to their performance on external datasets.

\subsection{Web Search on References Bias \label{sec:web_intro}}

As LLMs are black boxes for users, we utilized a Python-based automated web scraping tool to examine the presence of claims in \dsname on the web across 11 selected languages. This examination aims to identify potential biases in training datasets that cause imbalanced performance on the multilingual fact-checking task. 

To simulate varied internet user environments and bypass potential search engine restrictions, we incorporated diverse user agents, thus mirroring the wide spectrum of real-world internet access points. Further enhancing the authenticity of our approach, we introduced randomized time intervals between search queries, mimicking human browsing behavior and avoiding anti-bot mechanisms.

We performed web searches by Google's search engine to count the number of search results as a measure of the claim's online presence. This approach allows for a nuanced understanding of how widely each claim is disseminated across different linguistic contexts on the web. The detailed analysis is in section ~\ref{sec:web_exp}.

\subsection{Evaluating External Knowledge for Hallucination Reduction}

We employed the RAG technique, leveraging its state-of-the-art capabilities to bolster the accuracy and relevance of responses produced by LLMs within our fact-checking framework. This system applies the Dense Passage Retrieval (DPR) mechanism~\cite{karpukhin2020dense}, which utilizes embeddings for document retrieval. RAG revolutionizes NLP by amalgamating generative models with an external knowledge retrieval component, enabling dynamic access to a vast corpus of information. 
This external augmentation enhances the model's internal knowledge base with pertinent external data during generation.

For external retrieval, we employ the wiki\_dpr dataset, an extensive collection of 21 million Wikipedia passages, each adorned with DPR embeddings. These documents are segmented into 100-word, non-overlapping text blocks, optimizing the dataset for precise analysis and the evaluation of DPR's retrieval efficacy.

Leveraging Facebook AI Similarity Search (FAISS), we established an indexing framework based on the dataset's embeddings, streamlining the semantic retrieval of documents. Through the DPR Question Encoder, claims are transformed into semantically enriched embeddings. When these embeddings are matched against the FAISS index, the system identifies and retrieves the top 5 documents most relevant to the given claim. This retrieval process ensures the selection of documents that are semantically aligned with the claim. Consequently, the LLM is equipped with a rich input context that includes the original claim, constraints, and the substance of the retrieved documents.

\section{Evaluation and Analytics}

\subsection{Experimental Setup}

Our selection of ChatGPT, LLaMA-2, and LLaMA-3.1 was driven by their extensive language support and significant influence in the AI field. Although LLaMA-2 is primarily designed for English, it includes 27 other languages~\cite{touvron2023llama}, prompting us to examine its non-English hallucination detection. Initial evaluations (see Table~\ref{tab:compare_all}: columns 4, 5, 6) revealed its multilingual limitations, as its performance approximated random guessing. Therefore, we extended our evaluation to LLaMA-3.1, which introduces improvements in multilingual capabilities and factual consistency. This comparison allows us to assess whether the latest iteration reduces hallucinations in non-English settings.

We conduct evaluations of ChatGPT 3.5 Turbo, LLaMA-2 (7B, 13B, and 70B), and LLaMA-3.1 8B on a server equipped with a 12-core CPU and dual NVIDIA A5500 GPUs, each with 24GB of memory. For the evaluation of the larger LLaMA-2 70B model, we use a more powerful machine comprising 2 NUMA nodes. Each of these nodes features a 20-core CPU and 4 NVIDIA V100 GPUs with 32GB of memory, interconnected via NVLink to ensure fast and efficient data transfer. The LLaMA-3.1 8B model is evaluated on the same hardware as LLaMA-2 7B and 13B, ensuring a consistent computational environment for comparison. All models are evaluated using the PyTorch deep learning framework, version 2.0.1.
Furthermore, the temperature of LLaMA-2 is set to 0, which is intended to yield deterministic results. 

To further evaluate the performance of LLaMA 3.1 8B model, we conducted experiments on the TACC Lonestar6 system equipped with three NVIDIA A100 GPUs, each with 40GB of HBM2 memory, alongside a 64-Core processor with 128 total cores and 256GB RAM. The model was evaluated across temperature values ranging from 0.1 to 2.0. The results, as shown in Table~\ref{tab:results-LLaMA3.1-8B}, indicate that languages such as English exhibit improved performance with higher temperatures, peaking around 2.0, while others, like Chinese and Amharic, show a decline in performance as the temperature increases. Languages such as Japanese and Bengali maintain relatively stable performance across temperatures, while others like Tamil and Korean exhibit fluctuations. Overall, the optimal temperature setting varies by language, with some benefiting from higher temperatures and others being more stable at lower ones, indicating that temperature tuning is critical for achieving the best results across languages.
\begin{table*}[htbp]
    \centering
    \scriptsize
    \setlength{\tabcolsep}{3pt}
    \begin{adjustbox}{width=\textwidth}
    \begin{tabular}{lcccccccccccccccccccc}
        \toprule
        \textbf{Lang.} & \textbf{0.1} & \textbf{0.2} & \textbf{0.3} & \textbf{0.4} & \textbf{0.5} & \textbf{0.6} & \textbf{0.7} & \textbf{0.8} & \textbf{0.9} & \textbf{1.0} & \textbf{1.1} & \textbf{1.2} & \textbf{1.3} & \textbf{1.4} & \textbf{1.5} & \textbf{1.6} & \textbf{1.7} & \textbf{1.8} & \textbf{1.9} & \textbf{2.0} \\
        \midrule
        \textbf{en} & 47.72 & {\ul 47.71} & 47.74 & 47.8  & 47.78 & 47.89 & 48.04 & 47.97 & 48.06 & 48.25 & 48.39 & 48.52 & 48.47 & 48.78 & 48.79 & 48.83 & 49.06 & 49.07 & \textbf{49.26} & 49.17 \\
        \textbf{zh-CN} & \textbf{55.79} & 55.66 & 55.53 & 55.43 & 55.31 & 55.28 & 54.88 & 54.88 & 54.67 & 54.56 & 54.52 & 54.28 & 54.00 & 54.02 & 54.02 & 53.85 & 53.52 & 53.47 & 53.24 & {\ul 52.95} \\
        \textbf{hi} & 46.72 & {\ul 46.70} & 46.73 & 46.98 & 46.96 & 47.22 & 47.58 & 47.64 & 47.65 & 47.86 & 48.00 & 48.23 & 48.23 & 48.69 & 48.68 & 48.54 & 48.51 & 48.43 & 48.63 & \textbf{48.84} \\
        \textbf{ar} & 48.41 & {\ul 48.38} & 48.49 & 48.64 & 48.66 & 49.16 & 49.06 & 49.18 & 49.28 & 49.23 & 49.62 & 49.60 & 49.70 & 49.62 & 49.92 & 49.66 & 49.97 & \textbf{50.07} & 49.99 & 49.86 \\
        \textbf{bn} & 42.71 & {\ul 42.85} & 43.21 & 43.52 & 44.03 & 44.58 & 45.28 & 45.40 & 45.47 & 46.27 & 46.28 & 46.69 & 46.99 & 47.27 & 47.20 & 47.04 & \textbf{47.56} & 47.52 & 47.46 & 47.53 \\
        \textbf{ja} & 56.08 & \textbf{56.10} & 55.96 & 55.99 & 55.96 & 55.95 & 55.61 & 55.62 & 55.54 & 55.40 & 55.37 & 55.25 & 55.23 & 55.23 & 55.05 & 55.16 & 54.70 & 54.65 & 54.41 & {\ul 54.29} \\
        \textbf{ko} & 52.97 & 52.86 & \textbf{52.99} & 52.78 & 52.82 & 52.82 & 52.78 & 52.78 & 52.73 & 52.58 & 52.63 & 52.22 & 52.41 & 52.52 & 52.18 & 51.96 & 52.28 & 51.97 & 51.88 & {\ul 51.84} \\
        \textbf{ta} & 55.87 & \textbf{55.91} & 55.71 & 55.25 & 54.94 & 54.63 & 53.87 & 53.72 & 53.72 & 53.09 & 53.00 & 52.45 & 52.34 & 52.06 & 51.76 & 51.62 & 51.43 & 50.70 & 50.49 & {\ul 50.35} \\
        \textbf{th} & 49.61 & {\ul 49.76} & 49.92 & 49.86 & 49.89 & 50.04 & 50.22 & 50.15 & 50.22 & 50.48 & 50.45 & 50.54 & 50.77 & 50.61 & 50.77 & 50.81 & 50.66 & \textbf{50.93} & 50.58 & 50.46 \\
        \textbf{ka} & 47.86 & {\ul 47.94} & 48.19 & 48.13 & 48.35 & 48.40 & 48.59 & 48.66 & 48.64 & 48.79 & 48.91 & \textbf{49.26} & 49.09 & 49.04 & 49.16 & 49.26 & 48.95 & 48.86 & 48.65 & 48.61 \\
        \textbf{am} & 53.75 & \textbf{53.43} & 52.69 & 51.80 & 51.26 & 50.43 & 49.76 & 49.32 & 49.06 & 47.73 & 47.12 & 47.02 & 46.52 & 46.16 & 46.39 & 46.35 & 46.22 & {\ul 46.09} & 46.20 & 46.48 \\
        \bottomrule
    \end{tabular}
    \end{adjustbox}
    \caption{Performance of LLaMA-3.1 8B across temperature values from 0.1 to 2.0 for 11 languages. The highest accuracy for each language is highlighted in bold, while the lowest is underlined. The results provide insights into how different temperature settings influence multilingual fact verification and hallucination patterns across diverse linguistic contexts.}

    \label{tab:results-LLaMA3.1-8B}

    \vspace{-0.6cm}
\end{table*}
\subsection{Hallucination Detection in Multilingual Fact Verification}

\begin{table}[htbp]
\centering
\setlength{\tabcolsep}{3pt} 
\begin{tabular}{ccccccc}
\toprule
\textbf{Lang.} & \textbf{GPT} & \textbf{GPT Self.} & \textbf{L. 7B} & \textbf{L. 13B} & \textbf{L. 70B} &\textbf{L3.1 8B}\\
\midrule
\textbf{en}       & \textbf{65.89\%}     & \textbf{61.88\%}             & \textbf{63.35\%}    & \textbf{64.27\%}     & \textbf{64.56\%}  & 48.43\%     \\
\textbf{zh-CN}    & 58.25\%              & 53.94\%                      & 58.29\%             & 59.88\%              & 38.75\%    & 54.51\%           \\
\textbf{hi}       & 52.90\%              & 58.10\%                      & 48.59\%             & 54.33\%              & 45.68\%    & 47.94\%      \\
\textbf{ar}       & {\ul 45.48\%}        & 58.80\%                      & 50.55\%             & 54.97\%              & 32.97\%    & 49.33\%          \\
\textbf{bn}       & 55.65\%              & 58.73\%                      & 49.05\%             & 52.30\%              & 33.87\%    & {\ul45.86\%}          \\
\textbf{ja}       & 55.89\%              & 58.95\%                      & 58.24\%             & 59.57\%              & 41.37\%    & \textbf{55.34\%}          \\
\textbf{ko}       & 57.29\%              & 60.14\%                      & 56.67\%             & 58.74\%              & 46.06\%     & 52.47\%         \\
\textbf{ta}       & 55.67\%              & 59.98\%                      & 49.54\%             & 50.86\%              & {\ul 19.47\%}    & 53.18\%    \\
\textbf{th}       & 56.82\%              & 52.04\%                      & 53.90\%             & 53.46\%              & 48.09\%       & 50.37\%       \\
\textbf{ka}       & 57.37\%              & 51.69\%                      & 47.39\%             & 53.45\%              & 46.15\%       & 48.71\%       \\
\textbf{am}       & 47.53\%              & {\ul 47.06\%}                & {\ul 43.42\%}       & {\ul 48.64\%}        & 26.34\%      & 48.37\%       \\
\bottomrule
\end{tabular}
\caption{Comparison on accuracy of hallucination detection on fact-checking task by ChatGPT 3.5, ChatGPT 3.5 Self-Detection, LLaMA-2 7B, 13B, 70B and LLaMA-3.1 8B. Highest bolded, lowest underlined.}
\vspace{-0.8cm}
\label{tab:compare_all}
\end{table}

We deployed the same fact-checking process on ChatGPT 3.5, LLaMA-2 series to compare the hallucination detection abilities on multilingual claims. Specifically, we observed the self-detection ability of ChatGPT 3.5 by prompting it to rephrase the original claims and identify the validity of the generated context. 
In Table~\ref{tab:compare_all}, English consistently shows the highest accuracy for all models. 
For other languages, accuracy rates around 50\%, which is comparable to the expected outcomes of random guesses in binary-answer scenarios. It is also interesting that LLaMA-2 70B (see Table~\ref{tab:results-LLaMA-70B}) outperforms ChatGPT 3.5 (see Table~\ref{tab:results-chatgpt}) in English Climate-FEVER and SciFact, yet it demonstrates inferior performance in the non-English versions of Climate-FEVER and SciFact. 

Moreover, LLMs vary in self-detection accuracy across languages. The ChatGPT 3.5 model and its self-detection show performance differences, with the variant excelling in languages like Hindi due to targeted training. However, in English, the original model's broader training base provides superior accuracy, highlighting the impact of training scope and data diversity on self-detection capabilities.

\subsection{Web Search Results: Relationship Between Data Availability and Hallucination Rates \label{sec:web_exp}}

\begin{figure}[h]
    \centering
    \includegraphics[width=0.44\textwidth]{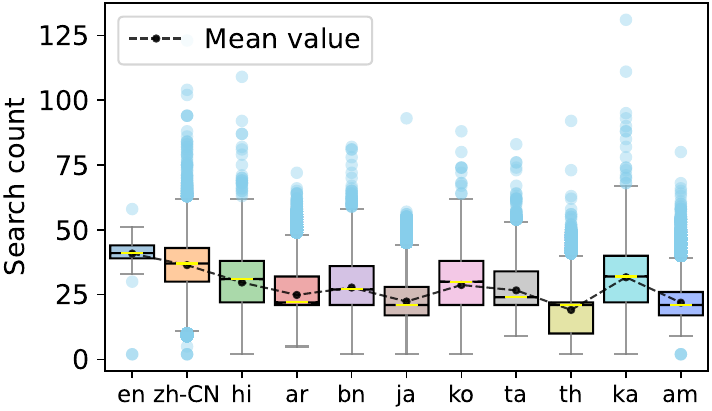}
    \caption{Web search distribution on multilingual claims. Middle 50\% of search counts inside each box, mean values for every language are connected.}
    \label{fig:web_search}
    \vspace{-0.4cm}
\end{figure}

We employed a Python-based automated web scraping tool to analyze the web presence of claims in 11 languages, aiming to detect biases in training datasets that could lead to uneven performance in multilingual fact-checking tasks, check it in the section ~\ref{sec:LDA_intro}.

Figure~\ref{fig:web_search} compares the count of search results across 11 different languages, with an emphasis on identifying potential biases in the training datasets. Some languages, like English, show a relatively wide interquartile range (IQR), which contains the middle 50\% of the data, indicating a high variability in the search count. Others, like Thai, have a much narrower IQR, indicating less variability. 
Languages like Amharic and Georgian have lower median and mean search counts, indicating less available content or fewer search results for these languages. This disparity could lead to an imbalanced performance in multilingual fact-checking, with better results in languages that have more content available, like English, and worse results in languages with less content.

\begin{figure}[h]
    \centering
    \includegraphics[width=0.44\textwidth]{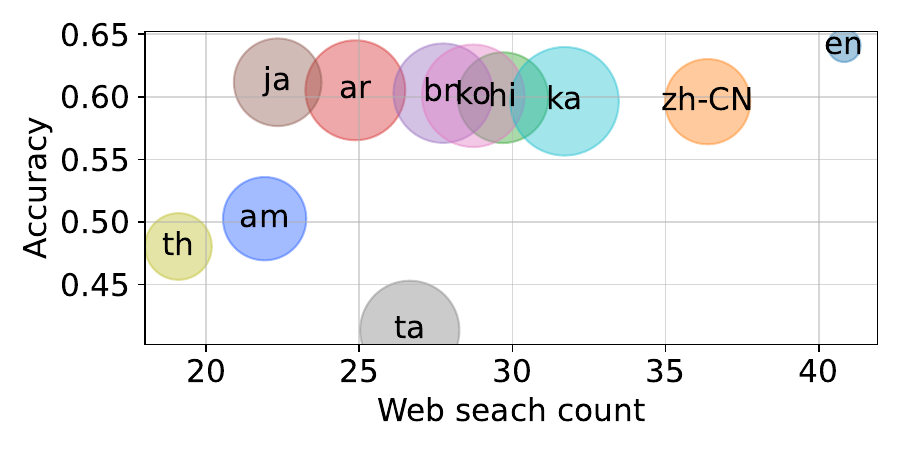}
    \caption{Detection accuracy on web search count. Bubble sizes depend on data variances to mean values on web search counts of each language.}
    \label{fig:web_acc}
    \vspace{-0.2cm}
\end{figure}

Figure~\ref{fig:web_acc} compares relationships over web search counts with hallucination detection accuracy in various languages. The analysis indicates a relevance between a language's web search frequency and its fact-checking accuracy, with English and Chinese demonstrating high levels on both counts. Lesser-known languages like Amharic and Tamil, with low web search frequencies, exhibit reduced accuracy, indicating that limited data negatively affects model learning. Furthermore, the data suggests potential biases, with models possibly favoring languages that dominate web content, affecting their accuracy in languages with less online presence.

\subsection{Topic-Level Analysis of Hallucination Detection \label{sec:LDA_exp}}

\begin{figure*}[h]
    \centering
    \includegraphics[width=\textwidth]{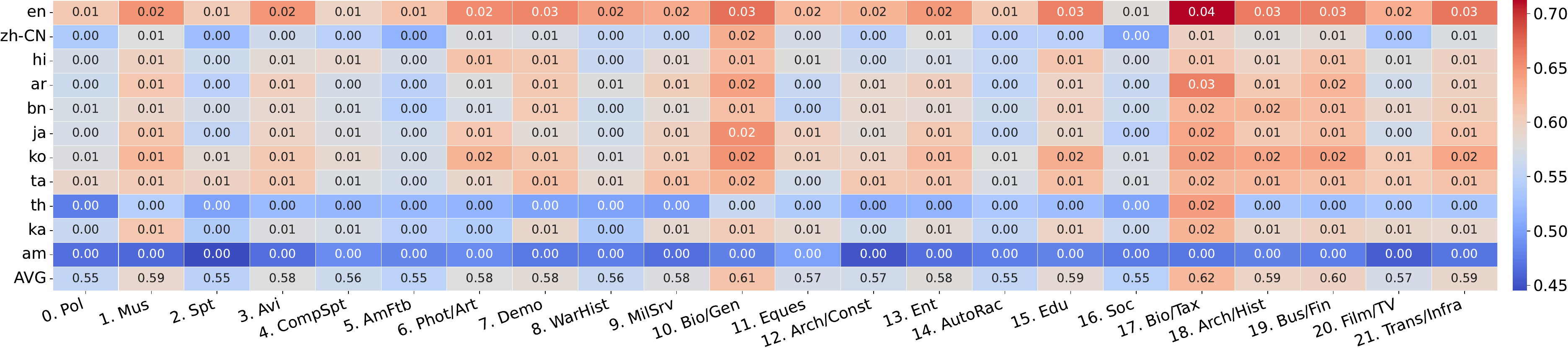}
    \vspace{-0.6cm}
    \caption{Average accuracy (represented by color) across topics by language, with standard deviation indicated by annotation. The color intensity reflects accuracy levels, where higher accuracy is represented by stronger shades, while lower accuracy is indicated by lighter tones. Standard deviation values illustrate the degree of variability within each topic, highlighting cases where the model exhibits inconsistent fact-checking performance across languages. The bottom row aggregates the mean accuracy and standard deviation for each topic across all languages, providing an overall assessment of topic difficulty and model robustness.}

    \label{fig:LDA}
    \vspace{-0.4cm}
\end{figure*}

\begin{table*}[htbp]
\centering
\resizebox{\textwidth}{!}{
\begin{tabular}{ccccccccccc}
\toprule
\textbf{Lang.} & \textbf{Poli} & \textbf{Sport} & \textbf{Comp} & \textbf{Football} & \textbf{WarHist} & \textbf{Equestr} & \textbf{ArchConst} & \textbf{AutoRace} & \textbf{Soccer} & \textbf{FilmTV} \\
\midrule
\textbf{zh-CN} & 2.62 & 2.5 & 1.17 & 4.72 & 2.78 & 3.02 & 1.02 & 1.28 & 2.66 & 2.59 \\
\textbf{hi}    & 1.97 & 0.14 & -1.33 & 2.80 & 4.66 & 1.87 & 0.88 & 0.62 & -0.39 & 1.52 \\
\textbf{ar}    & 1.57 & 0.42 & 0 & 4.99 & 0.91 & 1.47 & 0.73 & 1.99 & 0.29 & 1.64 \\
\textbf{bn}    & 0.86 & 0.28 & -0.81 & 3.31 & 2.16 & 3.75 & -0.58 & -0.66 & 0.09 & -1.13 \\
\textbf{ja}    & -0.35 & -1.81 & -1.38 & 1.11 & 1.21 & -2.85 & 0.22 & 0.43 & 0.35 & 0.21 \\
\textbf{ko}    & 0.4 & -2.36 & -0.76 & 1.08 & 1.87 & -0.98 & 0.22 & 0.66 & -0.83 & -0.32 \\
\textbf{ta}    & -1.41 & -4.31 & -0.48 & 1.05 & -1.62 & 0.33 & -0.95 & -1.89 & -1.98 & -3.00 \\
\textbf{th}    & 6.95 & 6.05 & 2.9 & 4.09 & 6.65 & 4.07 & 6.43 & 2.27 & 3.79 & 3.93 \\
\textbf{ka}    & 2.28 & 2.57 & -0.6 & 2.83 & 4.16 & 1.39 & 0 & 0.76 & 0.67 & 0.25 \\
\textbf{am}    & 7.71 & 7.44 & 3.76 & 5.05 & 6.98 & 4.24 & 5.41 & 3.60 & 4.41 & 7.09 \\
\bottomrule
\end{tabular}}
\caption{Percentage improvement of correct judgment on hallucination detection after translating nuanced topics into English. This table quantifies the effect of translating topic-specific claims into English before classification, measuring the relative accuracy improvement across 10 languages. Positive percentage values indicate an improvement in accuracy after translation, while negative values suggest a decline.}

\label{tab:improve-lda-combined}
\vspace{-0.6cm}
\end{table*}

\begin{table}[htbp]
\centering
\begin{tabular}{cccccc}
\toprule
\textbf{Lang.} & \textbf{Original} & \textbf{RevPrompt} & \textbf{LDA} & \textbf{LDA+RAG} \\
\midrule
\textbf{en}       & \textbf{65.89\%}  & \textbf{64.09\%}     & \textbf{63.48\%} & \textbf{60.5\%}         \\
\textbf{zh-CN}    & 58.25\%           & 59.61\%              & 57.81\%          & 53.74\%                  \\
\textbf{hi}       & 52.90\%           & 59.93\%              & 60.03\%          & 54.49\%                  \\
\textbf{ar}       & 45.48\%           & 60.52\%              & 59.71\%          & 55.67\%                  \\
\textbf{bn}       & 55.65\%           & 60.29\%              & 59.57\%          & 57.35\%                  \\
\textbf{ja}       & 55.89\%           & 61.16\%              & 58.44\%          & 55.44\%                  \\
\textbf{ko}       & 57.29\%           & 60.08\%              & 60.46\%          & 57.42\%                  \\
\textbf{ta}       & 55.67\%           & {\ul 41.34\%}        & 58.83\%          & 56.89\%                  \\
\textbf{th}       & 56.82\%           & 48.04\%              & 57.39\%          & 53.37\%            \\
\textbf{ka}       & 57.37\%           & 59.64\%              & 59.44\%          & 56.99\%                  \\
\textbf{am}       & {\ul 47.53\%}     & 50.26\%              & {\ul 54.24\%}    & {\ul 53.49\%}       \\      
\bottomrule
\end{tabular}
\caption{Accuracy of hallucination detection across different evaluation settings, including the original process, refined prompts, topic-aware evaluation using LDA, and retrieval-augmented context with LDA+RAG.}

\label{tab:all-methods}
\vspace{-1cm}
\end{table}

To investigate the induction of hallucinations, we utilized LDA to analyze detection performance over 22 topics in 11 languages, which is described in the section~\ref{sec:LDA_intro}. 

Figure~\ref{fig:LDA} highlights LLMs generally perform best in English and struggle with Amharic and Thai. Topics such as Politics, Sports, Film/Television, and Warfare History prove challenging across languages due to their subjective nature, where personal biases and interpretations can obscure the distinction between fact and opinion. The dynamic nature of these fields, coupled with the need for specialized knowledge in areas like Architecture/Construction and Competitive Sports, complicates fact-checking. Historical contexts in Warfare History and Automotive Racing add another layer of complexity, as historical records can be biased or incomplete. Emotional ties to topics like American Football and Film/Television can bias information, while the subjective interpretation of data in Sports or Business/Finance makes objective verification difficult. The absence of universal standards in evaluating greatness in sports or the arts further complicates claim verification.

The standard deviation reveals varying degrees of biased hallucinations across languages. While topics 6, 7, 9, 10, 13, 15, and 17 show high average accuracy, there is significant variance, with languages like English and Chinese exhibiting higher accuracy. This variance underscores how hallucination biases differ among languages, reflecting the complex interplay between linguistic context and the accuracy of LLM predictions on specific topics.

Using LDA for linguistic topic extraction, Table~\ref{tab:improve-lda-combined} presents the improvements in accuracy for non-English languages such as Chinese, Arabic, Thai, and Amharic across nuanced topics 0, 2, 4, 5, 8, 11, 12, 14, 16, and 20. The observed accuracy gains suggest that topic structuring provides additional contextual grounding, which helps compensate for weaker internal representations in lower-resource languages. However, this approach does not yield uniform benefits across all languages. Specifically, Tamil experiences a decline in performance across most of these nuanced topics, indicating that explicit topic awareness does not always align with the model’s existing representations or may interfere with its learned knowledge. These findings emphasize the varied impact of LDA-based topic structuring across languages, underscoring the need for language-adaptive methodologies that account for linguistic diversity and dataset-specific factors in hallucination detection.

\subsection{Effect of Topic Awareness and Retrieval Context in Hallucination Detection}

Beyond prompt design, structured contextual information can influence hallucination detection, particularly in multilingual settings where language resources and model training data vary. To examine this, we evaluate the impact of explicit topic awareness (LDA) and retrieval-augmented context (LDA+RAG) on factual claim classification. This analysis considers how topic metadata from LDA and external references from dense passage retrieval (DPR) affect model accuracy across languages.

Table~\ref{tab:all-methods} summarizes hallucination detection accuracy under different evaluation settings. The results indicate that topic structuring improves accuracy for lower-resource languages, while retrieval-based evaluation has mixed effects. Arabic and Amharic see the highest gains with LDA, increasing accuracy by 14.23\% and 6.71\%, respectively, suggesting that structured topic guidance helps compensate for weaker internal representations. In contrast, English and Chinese experience slight declines, implying that explicit topic structuring may interfere with the model’s existing topic awareness.

Retrieval-based evaluation (LDA+RAG) shows varied results. Arabic benefits the most, with accuracy increasing by 10.19\%, followed by Amharic at 5.96\%, suggesting that retrieved evidence aids languages with limited web resources. However, retrieval slightly reduces accuracy in Chinese and Japanese, possibly due to conflicts with the model’s internal knowledge.

The evaluation also highlights language-specific challenges. Some languages, such as Tamil and Thai, exhibit minor or negative shifts across different settings, indicating that tokenization, morphology, and dataset biases affect structured context processing. While topic structuring and retrieval augmentation benefit certain languages, they are not universally effective. The findings emphasize the need for adaptive evaluation frameworks that account for linguistic diversity and dataset properties.

\section{Conclusion}

This paper introduces {\dsname}, a publicly available multilingual fact extraction and verification benchmark containing 77,973 factual claims in 11 widely spoken languages for hallucination detection in generative language models. Our analysis of ChatGPT, LLaMA-2, and LLaMA-3 reveals the complexities of model performance under language-wise and classification prompts. By investigating hallucination patterns via LDA and automated web searches, we uncover how topic distribution and resource imbalance affect accuracy across languages.
Our experiments show that topic-aware evaluation (LDA) consistently benefits lower-resource languages, while RAG further improves performance when internal model knowledge is limited. However, these strategies are not universally effective, as high-resource languages with strong pre-trained representations sometimes see marginal or negative gains. These findings underscore the need for language-adaptive approaches, combining linguistic adjustments and resource-oriented strategies to enhance the reliability of multilingual fact verification. The proposed {\dsname} benchmark thus provides a foundation for future work in responsible, transparent, and robust large language model development.

\newpage
\section{Ethical Considerations}

The development of {\dsname} prioritizes ethical considerations, focusing on data diversity, fairness, and environmental impact. By incorporating multiple languages, particularly low-resource ones, {\dsname} promotes inclusivity and equitable representation in LLM research. Ensuring that multilingual benchmarks account for linguistic diversity is critical to preventing biases that disproportionately favor high-resource languages. Our goal is to provide a dataset that supports ethical research by fostering language diversity, improving accessibility, and encouraging responsible AI development.

Beyond dataset diversity, we recognize the broader ethical challenges in LLM evaluation. Biases in data collection, annotation, and curation can influence model predictions, leading to skewed performance across different demographic and linguistic groups. Researchers must assess such biases, particularly in factual verification tasks, to ensure fair evaluations and prevent unintended model behavior that may reinforce misinformation or systemic disparities.

Additionally, privacy and data provenance are crucial ethical concerns. All data sources used in {\dsname} originate from publicly available datasets, ensuring transparency and compliance with ethical guidelines. However, as LLMs increasingly interact with real-world information, addressing concerns around data usage, consent, and potential societal harm is necessary. We encourage future research on privacy-preserving methods and bias mitigation techniques to enhance fairness in multilingual benchmarks.

Finally, we highlight the need for continued interdisciplinary collaboration in AI ethics. Addressing fairness, bias, and accessibility requires input from linguists, ethicists, policymakers, and technologists. By openly sharing {\dsname} and its methodologies, we aim to foster transparency and collective efforts toward ethical AI development. Ongoing dialogue and evaluation will be essential to refining best practices and ensuring that multilingual AI systems are equitable, inclusive, and beneficial to all communities.

\section{Limitation}

While {\dsname} provides valuable insights into LLM hallucination detection across multiple languages, it has certain limitations. 
One key challenge is the variability in self-detection capabilities across different languages, which highlights the difficulty of standardizing hallucination detection across diverse linguistic contexts. Current LLMs exhibit inconsistent performance depending on language structure, training data availability, and model-specific biases.

To enhance multilingual hallucination detection, future work should explore language-specific training strategies that incorporate larger, more diverse datasets, especially for underrepresented languages. Additionally, improvements in benchmarking methodologies, such as standardized evaluation metrics for cross-linguistic performance, are necessary to establish more reliable comparisons. Further research should also consider how dataset biases and topic-specific hallucination tendencies impact model behavior across different domains.

By recognizing these limitations, we aim to encourage further refinement of multilingual hallucination detection methodologies, ensuring that {\dsname} serves as a foundation for more robust, inclusive, and ethically responsible AI evaluation frameworks.

\begin{acks}
This work was supported in part by the Microsoft Accelerate Foundation Models Research Grant and Google Cloud Research Credits.
\end{acks}


\vspace{-0.1cm}
\bibliographystyle{ACM-Reference-Format}
\bibliography{ref}

\appendix
\section{Appendix}
\label{sec:appendix}
\subsection{Hallucination on Multilingual Fact-checking Task}
As shown in Table~\ref{tab:results-chatgpt}, for ChatGPT 3.5, it is observed that the model demonstrates great stability across all topic fields (general, climate, and science facts), with English consistently showing the highest judgment accuracy and Arabic showing the lowest judgment accuracy. The performance gap between the highest and lowest percentages for the three datasets are 23.11\%, 38.22\%, and 35.06\%, indicating the greatest variability in the Climate-FEVER and the least in FEVER.

\begin{table}[htbp]
\centering

\begin{tabular}{cccc}
\toprule
\textbf{Lang.} & \textbf{FEVER} & \textbf{Climate-FEVER} & \textbf{SciFact} \\ \midrule
\textbf{en}    & {\textbf{65.89\%}}    & {\textbf{74.29\%}}             & {\textbf{71.57\%}}      \\
\textbf{zh-CN} & 58.25\%          & 55.12\%                   & 54.83\%            \\
\textbf{hi}    & 52.90\%           & 41.79\%                   & 42.14\%            \\
\textbf{ar}    & {\ul 45.48\%}    & {\ul 36.07\%}             & {\ul 36.51\%}      \\
\textbf{bn}    & 55.65\%          & 41.19\%                   & 43.15\%            \\
\textbf{ja}    & 55.89\%          & 55.00\%                      & 55.56\%            \\
\textbf{ko}    & 57.29\%          & 50.60\%                    & 54.11\%            \\
\textbf{ta}    & 55.67\%          & 55.95\%                   & 43.58\%            \\
\textbf{th}    & 56.82\%          & 46.31\%                   & 44.44\%            \\
\textbf{ka}    & 57.37\%          & 50.36\%                   & 49.93\%            \\
\textbf{am}    & {47.53\%}    & {39.52\%}             & {37.95\%} \\ 
\bottomrule
\end{tabular}
\caption{Accuracy of hallucination detection by ChatGPT 3.5 on {\dsname}. Max and min values in each column are highlighted in bold and underlined.}
\label{tab:results-chatgpt}
\end{table}

\begin{table}[]
\centering
\begin{tabular}{cccc}
\toprule
\textbf{Lang.} & \textbf{FEVER}   & \textbf{Climate-FEVER} & \textbf{SciFact} \\ \midrule
\textbf{en}    & {\textbf{63.35\%}} & {\textbf{77.70\%}}         & {\textbf{70.71\%}} \\
\textbf{zh-CN} & 58.29\%          & 58.59\%                 & 62.46\%          \\
\textbf{hi}    & 48.59\%          & 41.92\%                 & 40.33\%          \\
\textbf{ar}    & 50.55\%          & 46.53\%                 & 51.52\%          \\
\textbf{bn}    & 49.05\%          & 46.31\%                 & {\ul 36.04\%}    \\
\textbf{ja}    & 58.24\%          & 61.23\%                 & 64.54\%          \\
\textbf{ko}    & 56.67\%          & 52.36\%                 & 63.87\%          \\
\textbf{ta}    & 49.54\%          & 46.56\%                 & 35.03\%          \\
\textbf{th}    & 53.90\%           & 44.73\%                 & 48.49\%          \\
\textbf{ka}    & 47.39\%          & {\ul 43.00\%}              & 37.31\%          \\
\textbf{am}    & {\ul 43.42\%}    & {48.15\%}                 & {45.26\%}          \\    
\bottomrule
\end{tabular}
\caption{Accuracy of hallucination detection by LLaMA-2 7B on {\dsname}.}
\label{tab:results-LLaMA-7b}
\end{table}

\begin{table}[]
\centering
\begin{tabular}{cccc}
\toprule
\textbf{Lang.} & \textbf{FEVER} & \textbf{Climate-FEVER} & \textbf{SciFact} \\ \midrule
\textbf{en}    & {\textbf{64.27\%}}  & {\textbf{72.00\%}}              & {\textbf{72.44\%}}    \\
\textbf{zh-CN} & 59.88\%        & 62.82\%                 & 66.57\%          \\
\textbf{hi}    & 54.33\%        & 56.06\%                 & 57.33\%          \\
\textbf{ar}    & 54.97\%        & 56.77\%                 & 61.41\%          \\
\textbf{bn}    & 52.30\%        & 51.01\%                 & {\ul 52.25\%}    \\
\textbf{ja}    & 59.57\%        & 66.46\%                 & 67.65\%          \\
\textbf{ko}    & 58.74\%        & 62.97\%                 & 67.36\%          \\
\textbf{ta}    & 50.86\%  & 50.38\%                 & 61.02\%          \\
\textbf{th}    & 53.46\%        & 45.57\%          & 58.79\%          \\
\textbf{ka}    & 53.45\%        & 55.00\%                    & 62.69\%          \\
\textbf{am}    & {\ul 48.64\%}  & {\ul 44.44\%}           & {53.68\%}  \\
\bottomrule
\end{tabular}
\caption{Accuracy of hallucination detection by LLaMA-2 13B on {\dsname}.}
\label{tab:results-LLaMA-13B}
\end{table}

\begin{table}[]
\centering
\begin{tabular}{cccc}
\toprule
\textbf{Lang.} & \textbf{FEVER} & \textbf{Climate-FEVER} & \textbf{SciFact} \\ \midrule
\textbf{en}    & {\textbf{64.56\%}}  & {\textbf{78.42\%}}              & {\textbf{75.32\%}}    \\
\textbf{zh-CN} & 38.75\%        & 49.72\%                 & 42.08\%          \\
\textbf{hi}    & 45.68\%        & 45.96\%                 & 33.67\%          \\
\textbf{ar}    & 32.97\%        & 30.03\%                 & 31.92\%          \\
\textbf{bn}    & 33.87\%        & 19.46\%                 & 18.02\%    \\
\textbf{ja}    & 41.37\%        & 51.27\%                 & 42.77\%          \\
\textbf{ko}    & 46.06\%        & 47.88\%                 & 51.31\%          \\
\textbf{ta}    & {\ul 19.47\%}  & 19.85\%                 & {\ul 10.17\% }         \\
\textbf{th}    & 48.09\%        & 31.22\%          & 30.65\%          \\
\textbf{ka}    & 46.15\%        & 40.00\%                    & 33.03\%          \\
\textbf{am}    & 26.34\%  & {\ul 11.11\%}           & 22.11\%  \\
\bottomrule
\end{tabular}
\caption{Accuracy of hallucination detection by LLaMA-2 70B on {\dsname} data.}
\label{tab:results-LLaMA-70B}
\end{table}

LLaMA-2 (7B, 13B, and 70B) results are illustrated in Table~\ref{tab:results-LLaMA-7b}, \ref{tab:results-LLaMA-13B}, and \ref{tab:results-LLaMA-70B} respectively.
As the size of the LLaMA\-2 models increases, a noticeable bias towards different languages in LLMs becomes apparent. In the case of LLaMA-2 7B and 13B, Amharic exhibits the poorest performance in general fields, whereas Bengali demonstrates the weakest performance in categorizing science claims. LLaMA\-2 70B displays significant variation in performance across different languages, particularly in terms of the lowest accuracy. In general and science topics, Tamil exhibits remarkably low estimation rates, registering only 15.00\% and 10.17\%, respectively.

\end{document}